\documentclass[10pt,twocolumn,letterpaper]{article}

\usepackage{cvpr}              
\usepackage[accsupp]{axessibility}
\definecolor{cvprblue}{rgb}{0.21,0.49,0.74}
\usepackage[pagebackref,breaklinks,colorlinks,allcolors=cvprblue]{hyperref}

\def\confName{CVPR}
\def\confYear{2026}

\title{Distance-aware Soft Prompt Guidance for Multimodal Valence-Arousal Estimation}

\author{
Byeongjin Jung \quad 
Chanyeong Park \quad 
Sejoon Lim\thanks{Corresponding author} \\
Kookmin University, Republic Of Korea \\
{\tt\small \{wjdqudwls05, park4134, lim\}@kookmin.ac.kr}
}

\begin{document}
\maketitle
\begin{abstract}
Valence-arousal (VA) estimation is crucial for capturing the nuanced nature of human emotions in naturalistic environments. While pre-trained Vision-Language models like CLIP have shown remarkable semantic alignment capabilities, their application in continuous regression tasks is often limited by the discrete nature of text prompts. In this paper, we propose a novel multimodal framework for VA estimation that introduces \textit{Distance-aware Soft Prompt Guidance} to bridge the gap between semantic space and continuous dimensions. Specifically, we partition the VA space into multiple discrete regions, each associated with distinct textual descriptions. Rather than a hard categorization, we employ a Gaussian kernel to compute soft labels based on the Euclidean distance between the ground truth coordinates and the region centers, allowing the model to learn fine-grained emotional transitions. For multimodal integration, our architecture utilizes a CLIP image encoder and an Audio Spectrogram Transformer (AST) to extract robust spatial and acoustic features. These features are temporally modeled via Gated Recurrent Units (GRUs) and integrated through a hierarchical fusion scheme that sequentially combines cross-modal attention for alignment and gated fusion for adaptive refinement. Experimental results on the Aff-Wild2 dataset show that our semantic-guided approach improves performance over the official baseline and demonstrates robust performance on in-the-wild data.
\end{abstract}    
\section{Introduction}

Accurate emotion recognition is a fundamental component of effective human-computer interaction and psychological research. Among various representations, the valence-arousal (VA) space has gained significant attention as it captures the continuous and nuanced nature of emotional states, where valence represents the degree of positivity or negativity, and arousal indicates the level of excitement or calmness. Despite recent progress, estimating VA in-the-wild remains challenging due to unconstrained environmental factors such as varying lighting, occlusions, and diverse facial expressions.\cite{affectnet}

To address these challenges, multi-modal fusion strategies integrating visual and audio information have become a standard approach. Recent studies have successfully used deep neural networks, such as ResNet\cite{resnet} for spatial features and VGGish\cite{vggish} or Log-Mel spectrograms\cite{logmel} for acoustic cues, often combined with temporal modeling such as Temporal Convolutional Networks (TCNs)\cite{tcn} or Recurrent Neural Networks (RNNs)\cite{rnn} to capture dynamic changes. However, most existing methods treat VA estimation as a pure numerical regression problem, often overlooking the rich semantic information inherent in emotional expressions.

To better capture such semantic information, recent research has explored the use of large-scale vision-language models that encode semantic relationships between visual and textual representations. In particular, the emergence of Vision-Language Models (VLMs), such as CLIP\cite{clip}, offers a promising direction to incorporate semantic knowledge into affective computing. However, a significant gap exists between the discrete nature of text prompts in CLIP\cite{clip} and the continuous dimensions of the VA space. Directly applying CLIP\cite{clip} to regression tasks is non-trivial, as it requires a mechanism to align numerical coordinates with semantic descriptions.

In this paper, we propose a novel multi-modal framework that introduces Distance-aware Soft Prompt Guidance to bridge this gap. Our motivation stems from the observation that human emotional expressions are often described using a combination of semantic categories and intensities. To incorporate this into a regression framework, we partition the continuous VA plane into a $ 3 \times 3 $ grid, defining nine distinct emotional regions. Instead of assigning a single hard label, we utilize a Gaussian kernel to generate soft target probabilities based on the Euclidean distance between the ground truth coordinates and the region centers. This approach allows the model to learn the semantic transitions between different emotional states while maintaining the precision required for continuous estimation.

The main contributions of this work are summarized as follows:
\begin{itemize}
    \item We propose a Distance-aware Soft Prompting method that reinterprets VA regression as a semantic alignment task, enabling the effective use of CLIP's\cite{clip} pre-trained knowledge.
    \item We define 9 emotional regions and employ Gaussian soft-labeling to capture the fine-grained nuances and intensities of emotional states.
    \item We develop a robust multimodal architecture that combines CLIP\cite{clip} (Vision) and AST (Audio)\cite{ast} with hierarchical fusion through cross-modal attention and gated mechanisms.
    \item Our approach achieves competitive performance on the Aff-Wild2 dataset\cite{affwild2}, demonstrating the effectiveness of semantic-guided learning for VA estimation in-the-wild.
\end{itemize}
\section{Related Work}

\subsection{Valence-Arousal Estimation}

Valence–Arousal (VA) estimation is a fundamental framework in affective computing for modeling continuous emotional states, based on Russell’s circumplex model\cite{va1}. Unlike discrete emotion classification, VA representation enables modeling of fine-grained emotional transitions in a continuous space.

In practice, VA estimation is formulated as a regression task, where the Concordance Correlation Coefficient (CCC)\cite{ccc} is widely adopted as the evaluation metric due to its ability to jointly measure correlation and agreement. Large-scale datasets such as Aff-Wild2\cite{affwild2} have significantly advanced the field by providing in-the-wild annotations, forming the basis of the ABAW challenges. In addition to video-based benchmarks, several large-scale datasets have further advanced VA estimation. AffectNet\cite{affectnet} provides large-scale facial images with continuous VA annotations, enabling effective pretraining, while RECOLA\cite{recola} supports multimodal affect modeling with audio, visual, and physiological signals. Other datasets such as AFEW-VA\cite{afewva} and SEWA\cite{sewa} extend these efforts to more diverse and real-world scenarios.

Recent approaches in ABAW competitions have consistently adopted deep multimodal architectures that combine spatial feature extraction and temporal modeling. Early works leveraged CNN-based visual encoders and recurrent networks for temporal dynamics, such as the $M^3$T framework, which integrates 3D CNNs with bidirectional RNNs for joint audio-visual modeling\cite{m3f}. Subsequent methods introduced more advanced temporal modeling, including TCN and Transformer-based architectures to capture long-range dependencies in emotional signals\cite{tcntransformer}.

More recent ABAW solutions further emphasize multimodal integration and temporal interaction. For example, the winning method of the 8th ABAW challenge employs ResNet-based visual features, VGGish\cite{vggish} audio features, and TCN-based temporal modeling combined with cross-modal attention for VA regression\cite{abaw8th1st}. These trends indicate that state-of-the-art VA estimation relies on the joint design of spatial encoding, temporal modeling, and multimodal fusion. However, most existing approaches treat VA estimation as a purely numerical regression problem, without explicitly incorporating semantically interpretable emotional structures.

\subsection{Multi-modal Fusion}

Multi-modal fusion is a key component in affective computing, as emotional states are expressed through heterogeneous and complementary signals such as facial expressions and speech. Rather than the choice of modalities, the main challenge lies in how to effectively integrate them. Existing approaches can be broadly categorized into tensor-based interaction modeling, gating-based adaptive fusion, and attention-based cross-modal interaction.

Early works focused on modeling high-order interactions across modalities. Tensor Fusion Network (TFN)\cite{tfn} captures joint representations via tensor outer products, while Low-rank Multimodal Fusion (LMF)\cite{lmf} improves efficiency through low-rank decomposition. These methods highlight the trade-off between expressive interaction modeling and computational complexity.

Another line of research introduces gating-based fusion to handle modality reliability. Gated Multimodal Units (GMU)\cite{gated_fusion} learn to adaptively weight modalities depending on their reliability, addressing challenges such as noise, occlusion, and missing signals. This enables the model to dynamically control the contribution of each modality.

More recently, attention-based fusion has become dominant, especially for sequential affective analysis. Cross-modal attention\cite{attention} allows one modality to selectively attend to another, effectively modeling inter-modal dependencies. MulT\cite{mult} further demonstrates that directional cross-modal attention\cite{attention} can handle unaligned multimodal sequences and capture long-range dependencies.

Recent studies suggest that attention alone may be insufficient under real-world conditions where modality inconsistency is common. To address this, inconsistency-aware fusion methods such as \cite{iaca} combine attention\cite{attention} with gating mechanisms to suppress unreliable cross-modal interactions, enabling more robust integration.

This evolution is also reflected in recent ABAW challenge systems, where competitive approaches adopt a unified pipeline: modality-specific feature extraction, temporal modeling (\eg TCN\cite{tcn} or RNN), and cross-modal interaction before regression. In particular, attention-based interaction combined with temporal encoding has become a standard design for modeling dynamic dependencies between audio and visual streams. Benchmarks such as Aff-Wild2\cite{affwild2} have further established this paradigm, with CCC\cite{ccc} as the primary evaluation metric.

Despite these advances, two key limitations remain. First, the complementarity between modalities is often assumed rather than guaranteed\cite{mult,iaca,baltruvsaitis2018multimodal}, leading to degraded performance under noise, occlusion, and temporal misalignment. Second, most methods focus on feature-level fusion followed by direct regression, without explicitly enforcing semantic consistency in the fused representation. This limitation motivates our approach, which extends conventional fusion by incorporating semantic alignment through text-guided supervision in the VA space.
\begin{figure*}[t]
\centering
\includegraphics[width=1.0\textwidth]{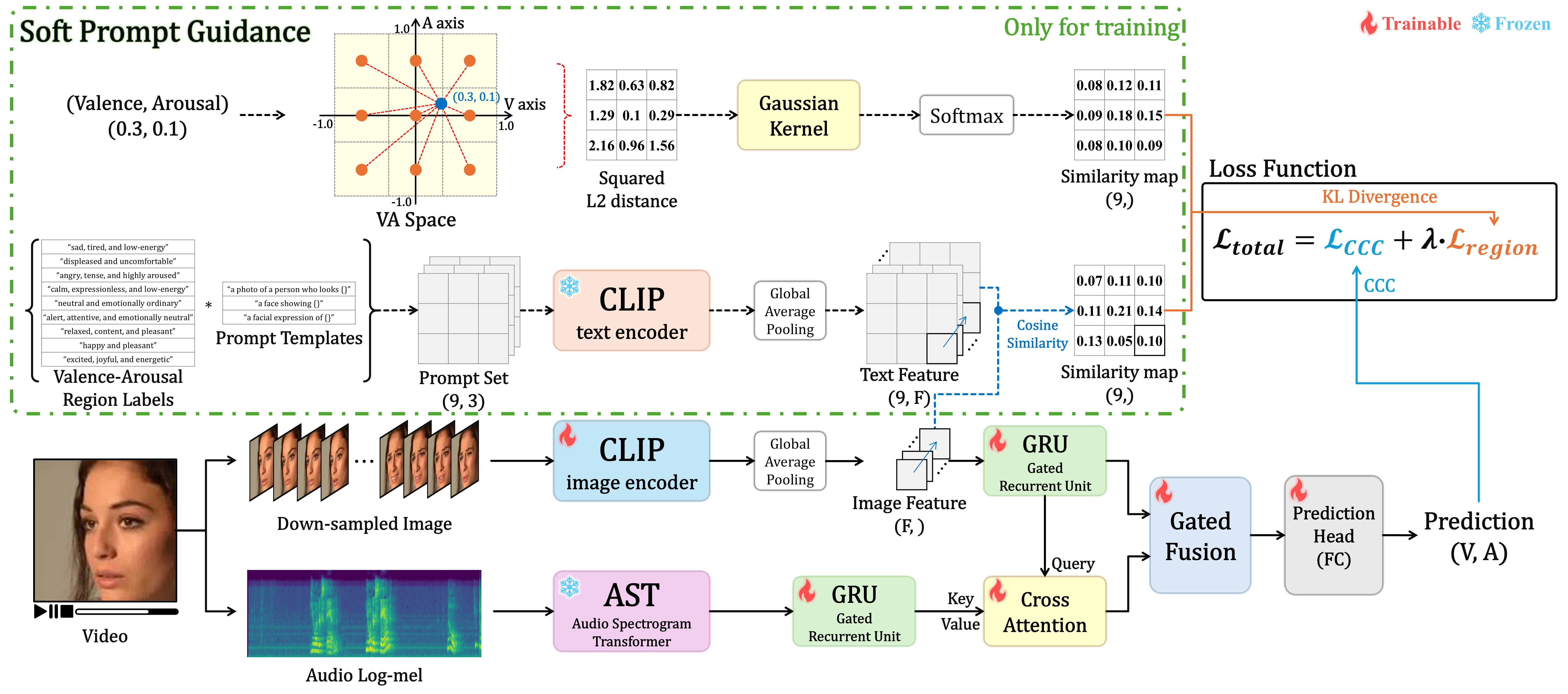}
\caption{\textbf{Framework Overview}: An overview of the proposed multimodal valence-arousal (VA) estimation framework. The framework extracts features from visual and audio inputs, and learns semantically enriched visual representations through soft prompt guidance. These features are then modeled temporally and fused in a multimodal manner to predict final VA values.}
\label{fig:framework_overview}
\end{figure*}

\subsection{Temporal Modeling}
Temporal modeling plays a crucial role in valence–arousal (VA) estimation because emotional expressions evolve continuously over time rather than appearing as isolated events. Facial expressions, speech prosody, and behavioral cues typically exhibit gradual transitions, making it essential to capture temporal dependencies when predicting continuous emotional states. Consequently, many affective computing systems incorporate temporal encoders to model the dynamics of emotional signals across sequential observations.

Early deep learning approaches primarily relied on recurrent neural networks (RNNs)\cite{rnn}, such as Long Short-Term Memory (LSTM)\cite{lstm} and Gated Recurrent Units (GRU)\cite{gru}, to model temporal dependencies in emotional signals. These architectures are capable of learning sequential patterns and have been widely used in emotion recognition tasks across modalities including speech, facial video, and physiological signals. For example, GRU\cite{gru}-based models have demonstrated strong capability in capturing temporal information in sequential emotional signals such as EEG or speech time-series data.

In addition to recurrent architectures, convolution-based temporal models have gained popularity due to their ability to capture long-range dependencies while maintaining efficient parallel computation. Temporal Convolutional Networks (TCNs)\cite{tcn} employ dilated causal convolutions to expand the receptive field and model long temporal contexts without relying on sequential recurrence. These properties make TCNs\cite{tcn} particularly suitable for continuous emotion recognition tasks where emotional states evolve gradually over extended time intervals.

More recently, transformer\cite{attention}-based architectures and hybrid temporal models have been introduced to further enhance the modeling of complex temporal dynamics. Several studies combine convolutional or recurrent encoders with transformer\cite{attention} modules to capture both local temporal patterns and long-range dependencies. For instance, recent work on continuous emotion recognition integrates TCN\cite{tcn} modules with transformer\cite{attention} encoders to learn spatiotemporal relationships across audio and visual streams, demonstrating improved performance on benchmarks such as Aff-Wild2\cite{affwild2}.

Another emerging direction focuses on modeling emotional dynamics across multiple temporal scales. Emotion signals often contain both short-term fluctuations and long-term trends, motivating multi-scale temporal representations. Recent approaches therefore incorporate hierarchical or multi-scale temporal modeling strategies to capture emotional variations occurring at different time horizons. Such designs allow the model to jointly represent instantaneous emotional responses and longer contextual trends, improving robustness in continuous emotion prediction.\cite{ye2023temporal}

Despite these advances, temporal modeling for VA estimation remains challenging. Emotional signals from different modalities are often noisy, partially observed, and asynchronously evolving, which complicates the modeling of consistent temporal dynamics. Furthermore, many existing approaches focus primarily on modeling temporal dependencies within each modality independently, while interactions between modalities and semantic emotional structures are addressed only at later stages of the architecture.\cite{baltruvsaitis2018multimodal}

In this work, we adopt GRU\cite{gru}-based temporal encoders to capture sequential emotional dynamics from both visual and acoustic feature streams. Compared with purely convolutional temporal encoders, recurrent architectures provide flexible modeling of temporal dependencies in variable-length sequences. The temporal representations are then integrated through cross-modal attention\cite{attention} and gated fusion\cite{gated_fusion} mechanisms, enabling the model to jointly capture temporal evolution and inter-modal interactions. 

\section{Proposed Method}

\subsection{Overview}

Our proposed framework consists of four main stages: (1) multi-modal feature extraction, (2) distance-aware soft prompt guidance, (3) temporal modeling, and (4) hierarchical multimodal fusion.

Given an input video segment, visual and audio features are first extracted using a CLIP image encoder\cite{clip} and an Audio Spectrogram Transformer (AST)\cite{ast}, respectively. During this stage, the distance-aware soft prompt guidance is applied to the visual branch, enabling the model to incorporate rich semantic knowledge into the feature representation.

Specifically, the proposed guidance mechanism bridges the gap between continuous VA coordinates and discrete emotional semantics by supervising the model to recognize the relative distances to predefined region prototypes. This allows the visual features to encode not only numerical VA information but also semantically meaningful emotional structures.

The extracted features are then temporally modeled and integrated through cross-modal attention\cite{attention} and gated fusion\cite{gated_fusion}, producing a unified representation for final valence-arousal prediction.

\subsection{Multi-modal Feature Extraction}

Our framework begins by extracting modality-specific representations from visual and audio inputs. For the visual branch, we employ a CLIP (ViT-B/16) image encoder\cite{clip} to obtain high-level visual features $F_v$ from facial image sequences. In addition, we leverage the CLIP text encoder\cite{clip} to construct semantic prototypes through Soft Prompt guidance, enabling the model to incorporate rich emotional semantics into the visual feature space.

Unlike conventional approaches that treat visual features as purely numerical representations, our design allows the visual encoder to capture both continuous VA-related information and semantically meaningful emotional cues. This is achieved by aligning visual features with text-based emotional descriptions derived from predefined VA regions.

For the audio branch, raw audio signals are transformed into Log-Mel spectrograms\cite{logmel} and processed using an Audio Spectrogram Transformer (AST)\cite{ast} to extract audio features $F_a$. These features capture complementary emotional information such as prosody and speech dynamics.

To facilitate multimodal interaction, both $F_v$ and $F_a$ are projected into a shared 256-dimensional embedding space through linear projection layers.

\subsection{Distance-aware Soft Prompt Guidance}

We propose a \textit{Distance-aware Soft Prompting} mechanism to bridge the gap between discrete semantic representations and the continuous valence-arousal (VA) space.

We partition the VA space into a $3 \times 3$ grid, resulting in $N=9$ emotion regions. Each region $i$ is associated with a semantic description (\eg ``low-energy'', ``uncomfortable''). To obtain robust textual representations, we employ three prompt templates and encode them using a CLIP text encoder (\eg ``a photo of a person who looks \{\}'', ``a face showing\{\}'', ``a facial expression of \{\}''). The region prototype $\mathbf{p}_i$ is defined as:
\begin{equation}
\mathbf{p}_i = \frac{1}{M} \sum_{m=1}^{M} \frac{f_{\text{text}}(t_{i,m})}{\|f_{\text{text}}(t_{i,m})\|}, \quad M=3
\end{equation}
where $t_{i,m}$ denotes the $m$-th prompt for region $i$.

Given an input image sequence, we extract frame-level features using the CLIP image encoder and compute a temporally aggregated representation:
\begin{equation}
\mathbf{v} = \frac{1}{T} \sum_{t=1}^{T} \frac{f_{\text{img}}(x_t)}{\|f_{\text{img}}(x_t)\|}
\end{equation}

We then measure the similarity between $\mathbf{v}$ and each region prototype $\mathbf{p}_i$ using scaled cosine similarity:
\begin{equation}
s_i = \tau \cdot \mathbf{v}^\top \mathbf{p}_i
\end{equation}
where $\tau$ is a temperature parameter. This similarity reflects the semantic consistency between visual features and region-level emotional descriptions, rather than purely numerical proximity in the VA space.

To incorporate continuous supervision, we convert the ground-truth VA coordinate $\mathbf{y} = (v, a)$ into a soft region distribution using a Gaussian kernel:
\begin{equation}
w_i = \frac{\exp\left(-\frac{\| \mathbf{y} - \mathbf{c}_i \|^2}{2\sigma^2}\right)}{\sum_{j=1}^{9} \exp\left(-\frac{\| \mathbf{y} - \mathbf{c}_j \|^2}{2\sigma^2}\right)}
\end{equation}
where $\mathbf{c}_i \in \{-0.66, 0.0, 0.66\}^2$ denotes the region centers and $\sigma$ is the smoothing parameter.

The model is trained with the semantic region loss $\mathcal{L}_{\text{region}}$ to align the predicted similarity distribution with the soft target distribution via KL divergence\cite{kl}:
\begin{equation}
\mathcal{L}_{\text{region}} = \text{KL}\big( \mathbf{w} \,\|\, \text{softmax}(\mathbf{s}) \big)
\end{equation}

This formulation enables smooth modeling of transitions between emotion states while leveraging semantically meaningful text prototypes as guidance, improving both representation learning and VA regression performance.

\subsection{Temporal Modeling and Hierarchical Multimodal Fusion}

To capture the dynamic nature of emotional expressions, we model temporal dependencies in both modalities using bidirectional Gated Recurrent Units (GRUs)\cite{gru}. Given visual and audio feature sequences, temporal representations are obtained as:
\begin{equation}
H_v = \mathrm{GRU}_v(F_v), \quad H_a = \mathrm{GRU}_a(F_a)
\end{equation}

To integrate multimodal information, we adopt a hierarchical fusion strategy. First, cross-modal attention\cite{attention} is applied to enrich visual features with audio features, where visual features act as the query and audio features serve as the key and value:
\begin{equation}
F_{\text{attn}} = \mathrm{Attention}(H_v, H_a, H_a)
\end{equation}

Next, we perform gated fusion\cite{gated_fusion} to adaptively combine the attended representation with the original visual context:
\begin{equation}
g = \sigma(W_g [F_{\text{attn}}; H_v] + b_g)
\end{equation}
\begin{equation}
F_{\text{fused}} = g \odot F_{\text{attn}} + (1 - g) \odot H_v
\end{equation}

This hierarchical design allows the model to first align cross-modal information and then selectively integrate it, preserving reliable visual cues while incorporating complementary audio information.

Finally, the fused representation is passed through a regression head to predict continuous valence and arousal values.

\subsection{Objective Function}

The model is optimized using a joint objective that combines regression accuracy and semantic alignment. The primary loss is the Concordance Correlation Coefficient (CCC) loss\cite{ccc}, denoted as $L_{\text{ccc}}$, which measures the agreement between the predicted and ground-truth valence and arousal values. Specifically, let $v_p, a_p$ and $v_g, a_g$ denote the predicted and ground-truth valence and arousal, respectively. The CCC loss $\mathcal{L}_{\text{ccc}}$ is computed as:

\begin{equation}
\mathcal{L}_{\text{ccc}} = 1 - \frac{1}{2} \left( \text{CCC}(v_p, v_g) + \text{CCC}(a_p, a_g) \right)
\end{equation}

In addition, we incorporate the region-level semantic alignment loss $L_{\text{region}}$ introduced in Section 3.3. 
The overall objective function is defined as:
\begin{equation}
\mathcal{L}_{\text{total}} = \mathcal{L}_{\text{ccc}} + \lambda \mathcal{L}_{\text{region}}
\end{equation}

where $\lambda$ is a weighting factor that balances regression performance and semantic consistency.

This joint optimization encourages the model to produce accurate VA predictions while maintaining semantically meaningful feature representations.
\section{Experiments and Results}

\subsection{Implementation Details}
The proposed framework is implemented using PyTorch and trained on a single NVIDIA GeForce RTX 3090 GPU. We utilize a batch size of 16 and train the model for 15 epochs. We employ the AdamW optimizer with a weight decay of $1 \times 10^{-4}$. To preserve the pre-trained semantic knowledge while allowing task-specific adaptation, we apply differential learning rates: $3 \times 10^{-6}$ for the backbones (CLIP ViT-B/16\cite{clip} and AST\cite{ast}) and $1 \times 10^{-4}$ for the task-specific heads. The hidden dimension for the GRU\cite{gru} and fusion modules is set to 256.

For data preparation, we utilize the Aff-Wild2 dataset\cite{affwild2} provided by the ABAW competition organizers. The dataset consists of approximately 500 videos captured in unconstrained ``in-the-wild'' scenarios, characterized by significant variations in lighting, head poses, and occlusions. We adopt a sliding window approach with a window size of $10$ seconds and a stride of $3$ seconds. From each $10$-second segment, $20$ visual frames are uniformly sampled at equal intervals to capture representative facial expressions. The corresponding audio is processed into a Log-Mel spectrogram\cite{logmel} to match the $10$-second window.

For evaluation, we follow a modified data split protocol. The official validation set of the Aff-Wild2 dataset\cite{affwild2} is used exclusively as the test set, ensuring that it is not involved in any stage of training. The original training set is further divided into training and validation subsets with an $8{:}2$ ratio for model development and hyperparameter tuning. All experimental results are reported on the validation subset constructed from the original training set. This setup guarantees that the reported results are obtained on unseen data without any information leakage from the evaluation set.

\subsection{Optimization of Temporal and Fusion Architectures}
We investigate the optimal architecture by comparing two temporal modeling types (GRU\cite{gru} and TCN\cite{tcn}) and three fusion strategies: Cross-modal Attention\cite{attention}, Gated Fusion\cite{gated_fusion}, and a hierarchical combination of both. As summarized in \cref{table:temporal_fusion}, the GRU-based temporal modeling combined with the hierarchical (Attn. + Gated) fusion achieves the best results. This suggests that while attention aligns the disparate modalities, the subsequent gating mechanism further refines the integration by adaptively weighting the most relevant information.

\begin{table}[ht]
\centering
\caption{Ablation study on temporal and fusion strategies.}
\label{table:temporal_fusion}
\resizebox{\columnwidth}{!}{%
\begin{tabular}{llccc}
\toprule
Temporal & Fusion & $CCC_v$ & $CCC_a$ & $CCC_{Mean}$ \\ \midrule
TCN & CM Attn. & 0.3068 & 0.5876 & 0.4472 \\
TCN & Gated & 0.3021 & 0.6017 & 0.4519 \\
TCN & Attn. + Gated & 0.4573 & 0.5679 & 0.5126 \\ \midrule
GRU & CM Attn. & 0.3623 & 0.6115 & 0.4869 \\
GRU & Gated & 0.3326 & 0.5657 & 0.4491 \\
\textbf{Ours (GRU)} & \textbf{Attn. + Gated} & \textbf{0.4655} & \textbf{0.6068} & \textbf{0.5361} \\ \bottomrule
\end{tabular}}
\end{table}

\subsection{Effect of Semantic Region Loss and CLIP Fine-tuning}
We investigate the impact of the semantic region loss weight $\lambda$ and the fine-tuning strategy of the CLIP\cite{clip} backbone. This experiment aims to determine the optimal balance between maintaining pre-trained semantic features and adapting to the VA regression task. \cref{tab:lambda_study} compares the performance of a completely frozen CLIP\cite{clip} backbone against a fine-tuned version with varying $\lambda$ values.

\begin{table}[ht]
\centering
\caption{Performance comparison across different $\lambda$ values and CLIP freezing strategies.}
\label{tab:lambda_study}
\resizebox{\columnwidth}{!}{%
\begin{tabular}{lcccc}
\toprule
Configuration & $\lambda$ & $CCC_v$ & $CCC_a$ & $CCC_{Mean}$ \\ \midrule
CLIP Frozen & - & 0.2896 & 0.5980 & 0.4438 \\
CLIP Fine-tuned & - & 0.3078 & 0.6318 & 0.4698 \\
CLIP Fine-tuned & 0.1 & 0.3070 & 0.6466 & 0.4768 \\
\textbf{Ours (CLIP Fine-tuned)} & 0.2 & \textbf{0.3520} & \textbf{0.6275} & \textbf{0.4897} \\
CLIP Fine-tuned & 0.3 & 0.3167 & 0.6585 & 0.4876 \\ \bottomrule
\end{tabular}}
\end{table}

\subsection{Sensitivity Analysis: Region Configuration}
Finally, we analyze the impact of the region center coordinates and the smoothing parameter $\sigma$. We compare the standard setting ($c \in \{-1.0, 0, 1.0\}, \sigma=0.6$) against a localized setting ($c \in \{-0.66, 0, 0.66\}, \sigma=0.45$). \cref{tab:grid_config} shows that the localized configuration provides superior performance. This is because the emotional intensities in the Aff-Wild2 dataset\cite{affwild2} are more frequently distributed within the $\pm 0.66$ range, allowing the soft prompting mechanism to provide more precise semantic guidance.

\begin{table}[ht]
\centering
\caption{Impact of grid center settings and $\sigma$ values.}
\label{tab:grid_config}
\resizebox{\columnwidth}{!}{%
\begin{tabular}{lcccc}
\toprule
Grid Centers ($c$) & $\sigma$ & $CCC_v$ & $CCC_a$ & $CCC_{Mean}$ \\ \midrule
$\{-1.0, 0.0, 1.0\}$ & 0.60 & 0.3656 & 0.5935 & 0.4795 \\
\textbf{Ours (\{-0.66, 0.0, 0.66\})} & \textbf{0.45} & \textbf{0.4655} & \textbf{0.6068} & \textbf{0.5361} \\ \bottomrule
\end{tabular}}
\end{table}

\subsection{Results}

We further analyze the experimental results presented in \cref{table:temporal_fusion}, \cref{tab:lambda_study} and \cref{tab:grid_config} to better understand the effectiveness of the proposed framework.

First, as shown in \cref{table:temporal_fusion}, the comparison of temporal modeling and fusion strategies demonstrates that the hierarchical fusion design consistently outperforms single-stage fusion methods. This can be attributed to the complementary roles of cross-modal attention and gated fusion. Cross-modal attention enables the model to capture interactions between visual and audio features, while the gating mechanism selectively controls the contribution of the attended features. This two-stage process allows the model to incorporate informative audio cues while suppressing unreliable or noisy signals, resulting in more robust multimodal representations.

Second, as shown in \cref{tab:lambda_study}, the effectiveness of the semantic region loss highlights the importance of incorporating semantic structure into VA estimation. Unlike conventional approaches that rely solely on numerical regression, the proposed method aligns visual features with region-level semantic prototypes, encouraging the model to learn semantically consistent representations. This alignment helps stabilize predictions and improves the coherence between estimated VA values and underlying emotional semantics.

Finally, as shown in \cref{tab:grid_config}, the sensitivity analysis of region configurations reveals that emotional expressions in in-the-wild data are more densely distributed around moderate intensity levels rather than extreme values. The localized region setting ({-0.66, 0, 0.66}) better reflects this distribution, enabling more precise semantic supervision. As a result, the model becomes more effective at capturing subtle emotional variations that are common in real-world scenarios.

Overall, these results suggest that the proposed framework benefits from the joint design of hierarchical multimodal fusion and semantic-guided learning, providing a more robust and semantically meaningful approach to VA estimation. These findings are consistent with the results on the ABAW competition test set, and the corresponding evaluation results are summarized in \cref{tab:ccc_comparison}.

\begin{table}[h]
\centering
\caption{Comparison of VA estimation performance on the Aff-Wild2 test dataset in terms of $CCC_{Mean}$.}
\label{tab:ccc_comparison}
\begin{tabular}{lc}
\toprule
Teams & $CCC_{Mean}$ \\ \midrule
RAS\cite{ras}  & 0.62 \\
EmoDX\cite{emodx}  & 0.58 \\
\textbf{Ours}  & 0.53 \\
HSEmotion\cite{hsemotion} & 0.52 \\ \midrule \midrule
Baseline & 0.22 \\ \bottomrule
\end{tabular}
\end{table}

In the ABAW competition, our method achieved 3rd place with a $CCC_{Mean}$ of 0.53. 
HSEmotion~\cite{hsemotion} focuses on visual feature extraction followed by regression, while EmoDX~\cite{emodx} emphasizes multimodal feature extraction and fusion. 
RAS~\cite{ras} further incorporates language features by injecting valence-arousal information into prompts and jointly leveraging visual, audio, and textual modalities. 
In contrast, our method utilizes a VLM to explicitly encode semantic priors in the continuous VA space into visual representations.
\section{Conclusion}

This paper presents a novel Valence-Arousal (VA) estimation framework that effectively integrates semantic knowledge through Distance-aware Soft Prompting. By applying a distance function based on the Gaussian kernel and precisely partitioning the VA space, we successfully bridged the gap between the continuous emotional space and the discrete nature of Vision-Language Models (VLMs). 

To optimize the model, we localized the semantic regions to centers at $\{-0.66, 0.0, 0.66\}$ and adjusted the smoothing parameter to $0.45$, thereby reflecting the natural distribution of emotional intensities. Furthermore, our hierarchical fusion scheme, which combines cross-modal attention and gated fusion, proved highly effective in capturing dynamic inter-modal relationships. Through extensive experiments, we investigated the impact of the semantic region loss weight $\lambda$; by utilizing it as a loss function, we identified the optimal balance between maintaining pre-trained features and adapting to the task, leading to significant performance improvements.

Leveraging these technical innovations, our model achieved outstanding results in the ABAW competition (Aff-Wild2 dataset)\cite{affwild2}, demonstrating its robustness and competitiveness in in-the-wild scenarios. Future research will focus on scaling this mechanism to diverse affective computing tasks and exploring context-aware prompt templates to further maximize the synergy between vision and language modalities.
\section{Acknowledgments}
This research was supported by the National Research Foundation of Korea (NRF) grant funded by the Korea government (MSIT) (RS-2026-25495200, Development of VLM-PINN-based End-to-End Autonomous Driving Technology Incorporating Vision-Language Understanding and Physical Consistency), and by the Korea Institute for Advancement of Technology (KIAT) grant funded by the Korea Government (MOTIE) (P0020536, HRD Program for Industrial Innovation).
{
    \small
    \bibliographystyle{ieeenat_fullname}
    \bibliography{main}
}


\end{document}